\documentclass[12pt]{article}
\usepackage[pdftex]{graphicx}
\usepackage{subcaption}
\usepackage{amsmath}
\usepackage{etoolbox}
\usepackage{array} 
\usepackage{ragged2e}
\usepackage[scientific-notation=true]{siunitx}
\usepackage{scicite}
\usepackage{times}
\usepackage{pdfpages}

\usepackage{hyperref}

\usepackage[utf8]{inputenc}

\newenvironment{sciabstract}{%
\begin{quote} \bf}
{\end{quote}}

\title{\vspace{-3.5cm}Secure and secret cooperation in robot swarms} 

\author{
Eduardo Castell\'{o} Ferrer*$^{1,2,3}$, 
Thomas Hardjono$^{2}$,\\ 
Alex `Sandy' Pentland$^{1,2}$,
Marco Dorigo$^{3}$\\
\small{$^1$MIT Media Lab,} \small{$^2$MIT Connection Science \& Engineering}\\ \small{Massachusetts Institute of Technology, Cambridge, MA 02139, USA}\\
\small{$^3$IRIDIA, Universit\'{e} Libre de Bruxelles, Brussels, Belgium}\\
\normalsize{$^\ast$E-mail:  ecstll@mit.edu.}}

\date{}

\begin{document} 

% Double-space the manuscript
\baselineskip 24pt

% Make the title.
\maketitle 

\begin{sciabstract}
The importance of swarm robotics systems in both academic research and real-world applications is steadily increasing. However, to reach widespread adoption, new models that ensure the secure cooperation of large groups of robots need to be developed. This work introduces a novel method to encapsulate cooperative robotic missions in an authenticated data structure known as Merkle tree. With this method, operators can provide the ``blueprint'' of the swarm's mission without disclosing its raw data. In other words, data verification can be separated from data itself. We propose a system where robots in a swarm, to cooperate towards mission completion, have to ``prove'' their integrity to their peers by exchanging cryptographic proofs. We show the implications of this approach for two different swarm robotics missions: foraging and maze formation. In both missions, swarm robots were able to cooperate and carry out sequential operations without having explicit knowledge about the mission's high-level objectives. The results presented in this work demonstrate the feasibility of using Merkle trees as a cooperation mechanism for swarm robotics systems in both simulation and real-robot experiments, which has implications for future decentralized robotics applications where security plays a crucial role such as environmental monitoring, infrastructure surveillance, and disaster management. 
\end{sciabstract}

%One-sentence summaries (containing no more than 135 characters and spaces) capturing the most important point should be submitted for all papers.
\section*{Summary}
\label{sec:Summary}
Robot swarms exploiting Merkle trees achieve secret and secure cooperation in sequential missions.

% The manuscript should start with a brief introduction that lays out the problem addressed by the research and describes the paper’s importance. The scientific question being investigated should be described in detail. The introduction should provide sufficient background information to make the article understandable to readers in other disciplines and provide enough context to ensure that the implications of the experimental findings are clear.
\section*{Introduction}
\label{sec:Intro}

Swarm robotics systems \cite{DorBirBra2014:sch-sr} have the potential to revolutionize many industries, from targeted material delivery \cite{Andrea2012} to precision farming \cite{Blender2016,Albani2017}. Boosted by technical breakthroughs, such as cloud computing \cite{Chen2010,He2016}, novel hardware design \cite{Majid2016,Mulgaonkar2018,rus2018design}, and manufacturing techniques \cite{CastelloFerrer2015}, swarms of robots are envisioned to play an important role in both industrial \cite{Chaudhari2016} and urban \cite{Starship:uspat,Alfeo2018urban} activities. The emergence of robot swarms has been acknowledged as one of the ten robotics grand challenges for the next 5-10 years that will have significant socioeconomic impact. Despite having such a promising future, many important aspects which need to be considered in realistic deployments are either underexplored or neglected \cite{Yang2018a}. 

One of the main reasons why swarms of robots have not been widely adopted in real-world applications is because there is no consensus on how to design swarm robotics systems that include perception, action, and communication \cite{Yang2018a}. In addition, recent research points out that the lack of security standards in the field is also hindering the adoption of this technology in data-sensitive areas (e.g., military, surveillance, public infrastructure) \cite{Scharre2014,vivek2019cyber}. These research gaps are motivating scientists to focus on new fields of study such as applied swarm security \cite{Akram2017,Gil2017} and privacy \cite{Prorok2016,Prorok2018} as well as to revisit already accepted assumptions in the field. 

\begin{figure*}[!t]
\begin{center}
\includegraphics[width=0.755\linewidth]{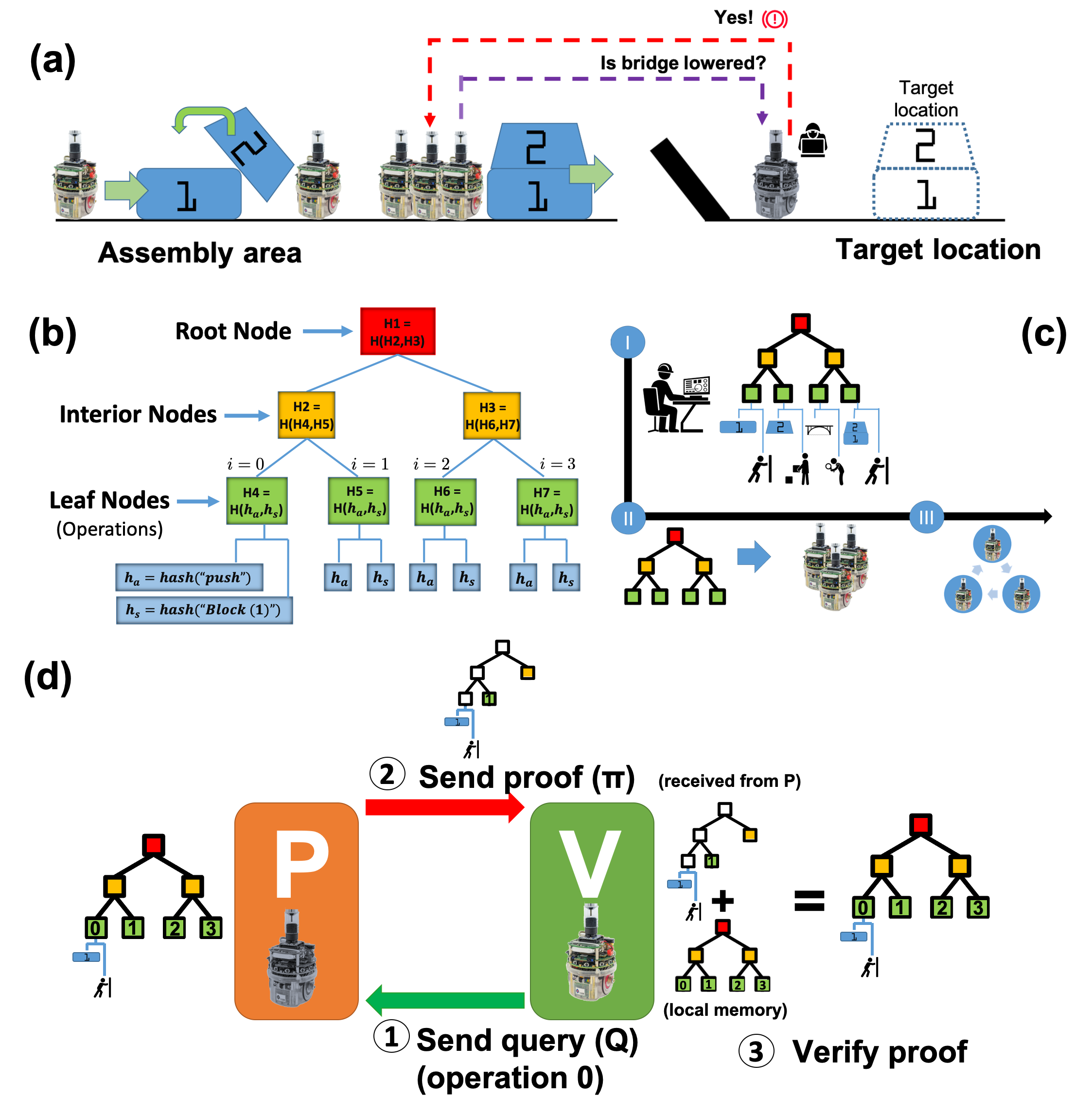}
\caption{\textbf{Towards secure and secret cooperation in swarm robotics missions.} (a) Example of sequential mission (e.g., build a 2-block tower and transport it to a target location) conducted by a robot swarm that can fail if one of the robots involved is tampered with. If a robot spreads erroneous information within the system (e.g., the robot in charge of making the bridge crossable sends a message confirming the bridge is lowered when it is not) the entire mission can fail. (b) Merkle tree (MT) used in this work: a hash-based tree structure where each leaf node stores the hash of an associated operation within the swarm's mission (i.e., the combination of a robot action hash ($h_{a}$) and a sensor input hash ($h_{s}$)), while each interior node contains the hash of the combination of its two children. (c) Mission initialization workflow: (I) The swarm's operator generates a valid MT where all the operations that the swarm needs to perform to complete its mission are included in the correct order. (II) The MT is broadcast to all the robots in the swarm. (III) The mission starts. (d) Workflow for \emph{prover} (P) and \emph{verifier} (V) robots when they exchange queries (Q) and MT ``proofs'' ($\pi$) to synchronize and complete their corresponding MTs copies.}
\label{fig:GeneralOverview}
\end{center}
\end{figure*}

From the origins of swarm robotics research, robot swarms were assumed to be fault-tolerant by design, due to the large number of robot units involved \cite{ChrOGrBirDor2007:ecal,ChrOGrDor2008:iros,NouGroBon-etal2009:tec,Millard2013}. However, it has been shown that a small number of partially failed (with defective sensors, broken actuators, noisy communications devices, etc.) \cite{Bjerknes2013} or malicious robots \cite{Strobel2018, Strobel2020} can have a significant impact on the overall system reliability and performance. The first surveys of swarm robotics security were presented in \cite{Winfield2006,Higgins2009}. These works identified physical capture and tampering with members as significant threats to robot swarms. Physical capture of a robot might not only lead to loss of availability but also to the capture of security credentials or critical details about the swarm operation \cite{Sargeant2016}. For instance, if a robot is tampered with and reintroduced into the swarm, an attacker might influence the behaviour of the whole system \cite{Millard2013,Strobel2018} and eventually hinder the entire mission. Fig.~\ref{fig:GeneralOverview}(a) depicts an example of such vulnerability, where four operations must be performed in a sequential order by a group of robots: first, block (1) needs to be pushed to an assembly area; second, block (2) has to be placed on top of (1) to assemble a tower-shaped structure; third, the state of a bridge that connects the assembly area and the target location needs to be determined as crossable (i.e., lowered down); fourth, the tower-shaped structure must be pushed to the target location at the other side of the bridge. If a robot is tampered with (e.g., the robot in charge of determining the status of the bridge), it can spread erroneous information within the system (e.g., tell the other robots that the bridge is crossable when it is not), therefore hindering the entire mission. This type of attack would be unique to swarm and multi-robot systems and is particularly critical in situations where robots must share data with other robots in the swarm or with human operators.

In previous swarm robotics work, researchers hard-coded the complete set of rules that trigger the transitions from operation to operation \cite{NouGroBon-etal2009:tec,Dorigo2013,Werfel2014} in all robots; that is, each robot in the swarm had a full copy of the information necessary to accomplish its mission. Although this distributed approach is more robust and fault-tolerant than centralized methods, it significantly increases the attack surface (i.e., the total sum of vulnerabilities) for an attacker to figure out the swarm's high-level goals and disrupt the system's behavior \cite{Sargeant2016}. Due to these concerns, in this work we aim to find an answer to the following questions: Is there a way to provide the ``blueprint'' of a robotic mission without disclosing the raw information describing the mission itself? In other words, is it possible for robots in a swarm to fulfill sequential missions without exposing knowledge about the mission's high-level objectives? To answer these questions, we present for the first time a cooperation model based on the idea of encapsulating robotic missions into Merkle trees. Our approach allows robots in a swarm to cooperate while minimizing security risks due to issues like physical capture or tampered members. We achieve this by creating a secure data object that is shared by the robots (i.e., a Merkle tree) instead of protecting the communication channel among them (e.g., by encrypting the network connections).

In this paper, we argue that by using Merkle trees (see Fig.~\ref{fig:GeneralOverview}(b), Materials and Methods, and Supplementary Materials), swarm operators can provide the ``blueprint'' of the swarm's objectives without disclosing raw or unprotected data about the mission itself (Fig.~\ref{fig:GeneralOverview}(c)). More specifically, we introduce for the first time a framework where data verification is separated from data itself. By exchanging cryptographic proofs, robots in the swarm are able to prove to their peers that they know specific pieces of information included in the swarm's mission and therefore that they are cooperating towards its completion (Fig.~\ref{fig:GeneralOverview}(d)). This approach was analyzed in simulation and real-robot experiments for two different sequential missions: foraging and maze formation. In both missions, robots were able to cooperate and carry out sequential operations without having explicit knowledge about the mission's high-level objectives. Our findings show that larger swarms tend to both increase the performance of the system and to diversify the amount of information within the swarm. However, larger numbers of robots as well as longer missions determine a linear increase in the communication requirements of the system. Nevertheless, an analysis of storage usage, communication costs, and computational time for larger-scale missions where the number of operations takes relatively large values reveals that the use of Merkle trees is within reach of current robotic technology.

% The results should describe the experiments performed and the findings observed. The Results section should be divided into subsections to delineate different experimental themes. Subheadings should either be all phrases or all complete sentences. They should be brief, ideally less than 10 words. Subheadings should not end in a period. Your paper may have as many subheadings as are necessary.

\section*{Results}
\label{sec:Results}

\subsection*{Foraging mission experiments}
\label{sec:Foraging}

\begin{figure}[tbh]
\begin{center}
\includegraphics[width=0.75\linewidth]{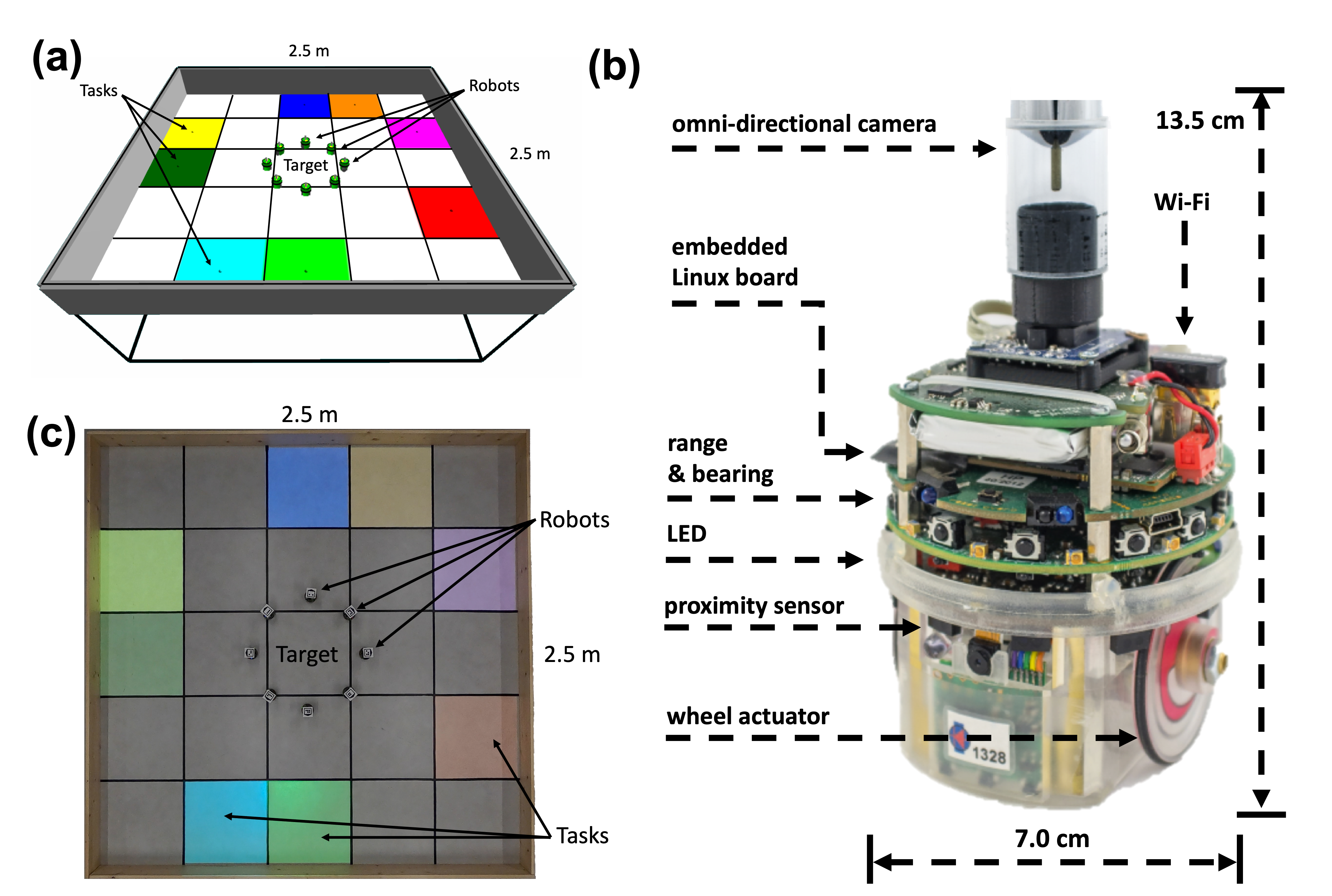}
\caption{\textbf{Foraging scenario}. (a) Simulated arena used for the foraging mission. The scenario consists of a rectangular area of 2.5 $\times$ 2.5 m$^{2}$ within which robots, tasks (represented as colored cells), and a target area (0.5$\times$0.5 m$^{2}$) located at the center are placed (see Supplementary Materials for a detailed description of the simulation experimental setup). A typical simulated run is displayed in \href{https://youtu.be/LGXWx_xMexw}{movie M1}. (b) Diagram of the e-puck robot together with its sensors layout. E-puck's size (13 cm tall and 7 cm in diameter) and features make it an ideal platform for simulated and real-world swarm robotics experiments. (c) Real-world equivalent of the scenario shown in (a). In this scenario, colored cells are projected on the arena by using a projector (see Supplementary Materials for a detailed description of the real-robot experimental setup). A typical real-world run is displayed in \href{https://youtu.be/hImae_ykp8M}{movie M2}.}

\label{fig:ForagingScenario}
\end{center}
\end{figure}

A set of 100 simulation experiments were carried out to analyze our approach in the foraging scenario shown in Fig.~\ref{fig:ForagingScenario}(a). The environment consists of a rectangular area segmented in a 5 $\times$ 5 grid with a central square representing the target location where e-puck robots \cite{Mondada2009:epuck} (Fig.~\ref{fig:ForagingScenario}(b)) need to deliver discovered tasks. In Fig.~\ref{fig:ForagingScenario}(a) tasks are represented by colored cells. Tasks have to be discovered by robots and then delivered to the target location following a predefined sequence of colors. The foraging mission is finished when all tasks in the sequence are delivered in the right order at the target location. In other words, when a robot in the swarm delivers the last task in the sequence. This sequence of colors is encoded by the swarm's operator as an MT (Fig.~\ref{fig:GeneralOverview}(b)) with $n$ operations (i.e., MT length) before the experiment starts (Fig.~\ref{fig:GeneralOverview}(c)). Therefore, during the mission, robots do not have explicit information about what is the correct sequence of colors to be delivered since they only have the encrypted information included in their MT. In addition, 10 real-robot experiments for a subset of the experimental conditions considered in the simulation experiments were carried out to validate the approach. Fig.~\ref{fig:ForagingScenario}(c) shows the real-robot foraging environment where tasks are represented by colored cells projected on the arena by using a projector (see Supplementary Materials for the experimental setup of the real-robot experiments). 

In the foraging mission, robots search the environment looking for tasks. Once a tasks is found, robots identify its color and generate $h_{s}$. Then, they generate the value $h_{a}$ for all possible actions.\footnote{In the considered foraging mission, $h_{a}$ is restricted to the ``carry to target'' action.} $H_{i}$ is calculated by hashing $h_{s}$ and $h_{a}$: $H_{i} = H((h_{s},h_{a}))$ and verified against the current working operation ($i$) of the robot; $i$ is initialized to 0 at the beginning of the mission and is an index that points to the operation that the robot is currently trying to fulfill in its local MT. In case it is possible to generate a valid proof (see Supplementary Materials for details about MT proofs) with the task's information ($\exists\pi_{i},H_{i}$), the robot ``grabs'' the task color and delivers it in the target location. If not, the robot keeps wandering. When an operation is completed successfully, the robot changes the status of $i$ as completed in its local MT. Then, the robot increases $i$: $i=i+1$ for $i \in (0 \leq i \leq n-1$). It is important to note that during the experiments, robots can exchange information with their peers (e.g., $i$, $\pi$). If two robots are within communication distance, they can exchange their $i$ values. If there is a disparity in the values (i.e., one $i$ is lower than the other) the robot with the lower $i$ will become a \emph{verifier} ($V$), and will send queries ($Q$) to the robot with greater $i$, the \emph{prover} ($P$), asking for proofs ($\pi$) about the missing operations as depicted in Fig.~\ref{fig:GeneralOverview}(d). This process is conducted until the $i$ value is the same for both robots. By using this method, robots can synchronize their own MT copies and therefore cooperate towards the fulfillment of the swarm's mission (see Supplementary Materials for robot's interaction and finite state machine diagrams). 

Next, we study how the performance of our approach (i.e., how fast and reliably a particular mission is carried out) changes when varying the MT length and the swarm sizes. We present results obtained in simulation and real-robot experiments where the MT length $n$ is varied in the set $[2,4,5,6,7,8]$ (simulation) and $[2,4,8]$ (real) and the swarm sizes $R_{n}$ in the interval $1 \leq R_{n} \leq 16$. In addition, the amount of communication (i.e., the total amount of data robots exchanged during the mission), and information diversity (i.e., how widely spread information is within the swarm) metrics are calculated and presented. Details about these analysis metrics and statistical methods used can be found in the Materials and Methods section. 
 
\includepdf[pages=1,scale=0.72,offset=0 110, pagecommand={\null\enlargethispage{2\baselineskip}\vfill\captionof{figure}{\textbf{Finishing times and probability of success for the foraging mission}. (a) Average $F_{t}$ (in seconds) and its standard deviations for different MT lengths ($n$) and robot swarm sizes ($R_{n}$) for both simulation (bars) and real-robot experiments (circles). (b) Probability of success ($P_{s}$) for all simulation configurations shown in (a) for the execution of the foraging mission with $R_{n}$ in the interval $1 \leq R_{n} \leq 16$. For each of the $R_{n}$ values a lower-bound (solid line representing $n=2$), mean of the evaluated interval (dot-dashed line), and the upper-bound (loosely dashed line representing $n=8$) were included. (c-e) Detailed plots for the simulation experiments depicting the whole $n$ set for (c) $R_{n}$ = 16, (d) $R_{n}$ = 4, (e) $R_{n}$ = 2. (f-g) Detailed plots for the real-robot experiments depicting the $n = [2,4,8]$ subset for (f) $R_{n}$ = 16, (g) $R_{n}$ = 4, (h) $R_{n}$ = 2. }\label{fig:Foraging}}]{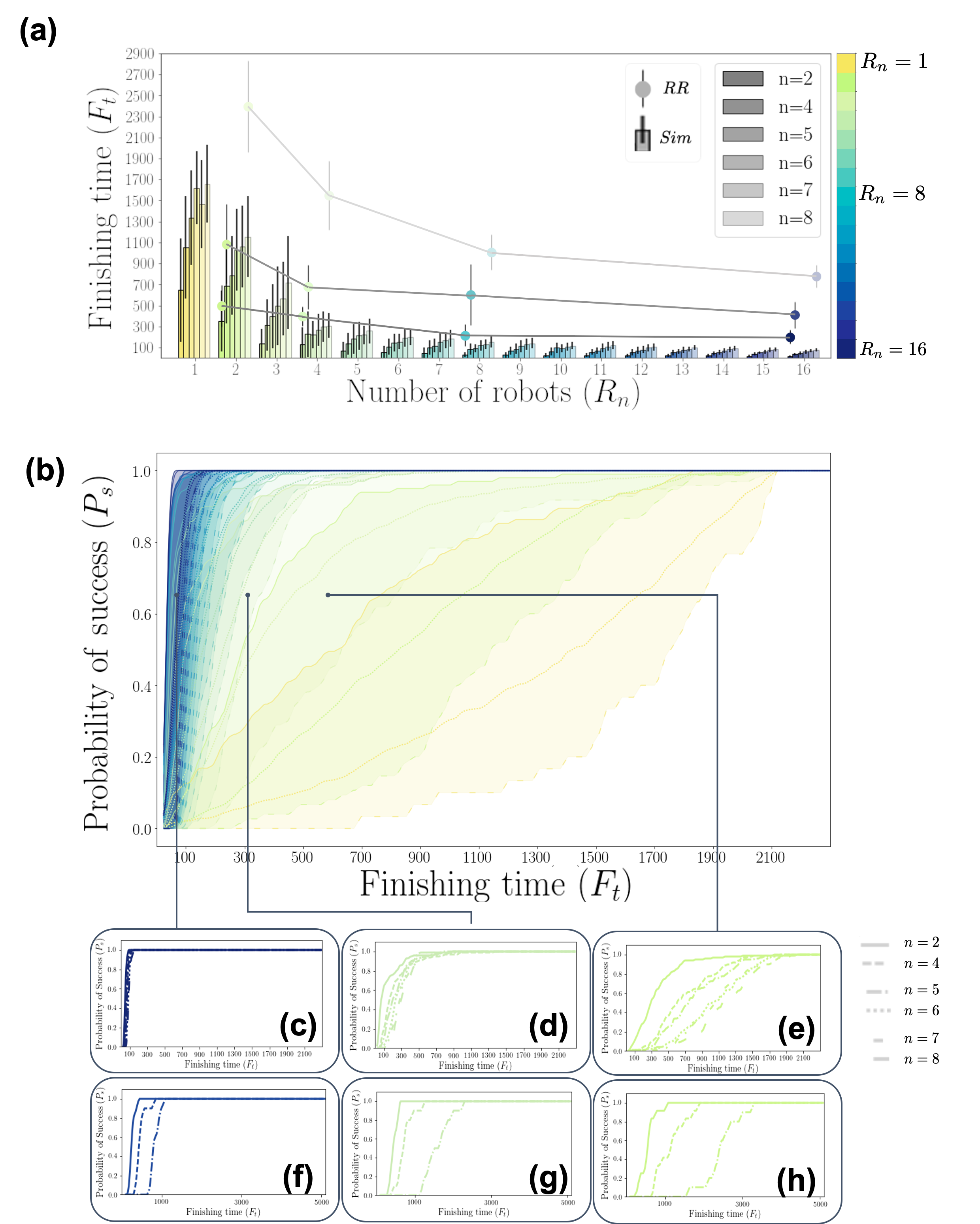}   

Fig.~$\ref{fig:Foraging}$(a) shows the finishing time ($F_{t}$) and its standard deviation for several MT length configurations ($n$) and robot swarm sizes ($R_{n}$) for both simulation and real-robot experiments. According to Fig.~\ref{fig:Foraging}(a), $F_{t}$ decreases when robots are added to the swarm, regardless of the length of the MT. These results also suggest that once a certain number of robots is present ($R_{n}\geq 8$), the length of the MT ($n$) has little impact on the $F_{t}$ of the system. Fig.~$\ref{fig:Foraging}$(b) shows how the probability of success ($P_{s}$) curves change when changing the number $n$ of tasks for all simulation configurations presented in Fig.~\ref{fig:Foraging}(a). One can also see that adding robots increases $P_{s}$ since the curves become steeper and converge to higher values sooner. However, these results also suggest that as we increase $n$ (the mission becomes longer), $P_{s}$ reaches higher values later (especially for $R_{n}\leq 3$). This effect can be seen in both simulation Fig.~\ref{fig:Foraging}(c-e) and real-robot experiments Fig.~\ref{fig:Foraging}(f-h). It is important to note that $P_{s}$ does reach 1 (the maximum value) in all configurations as experiments were run up to a Time Cap (TC), which was never reached (see Supplementary Materials for a detailed description of the experimental setup).

\begin{figure}[!t]
\centering
\includegraphics[width=\linewidth]{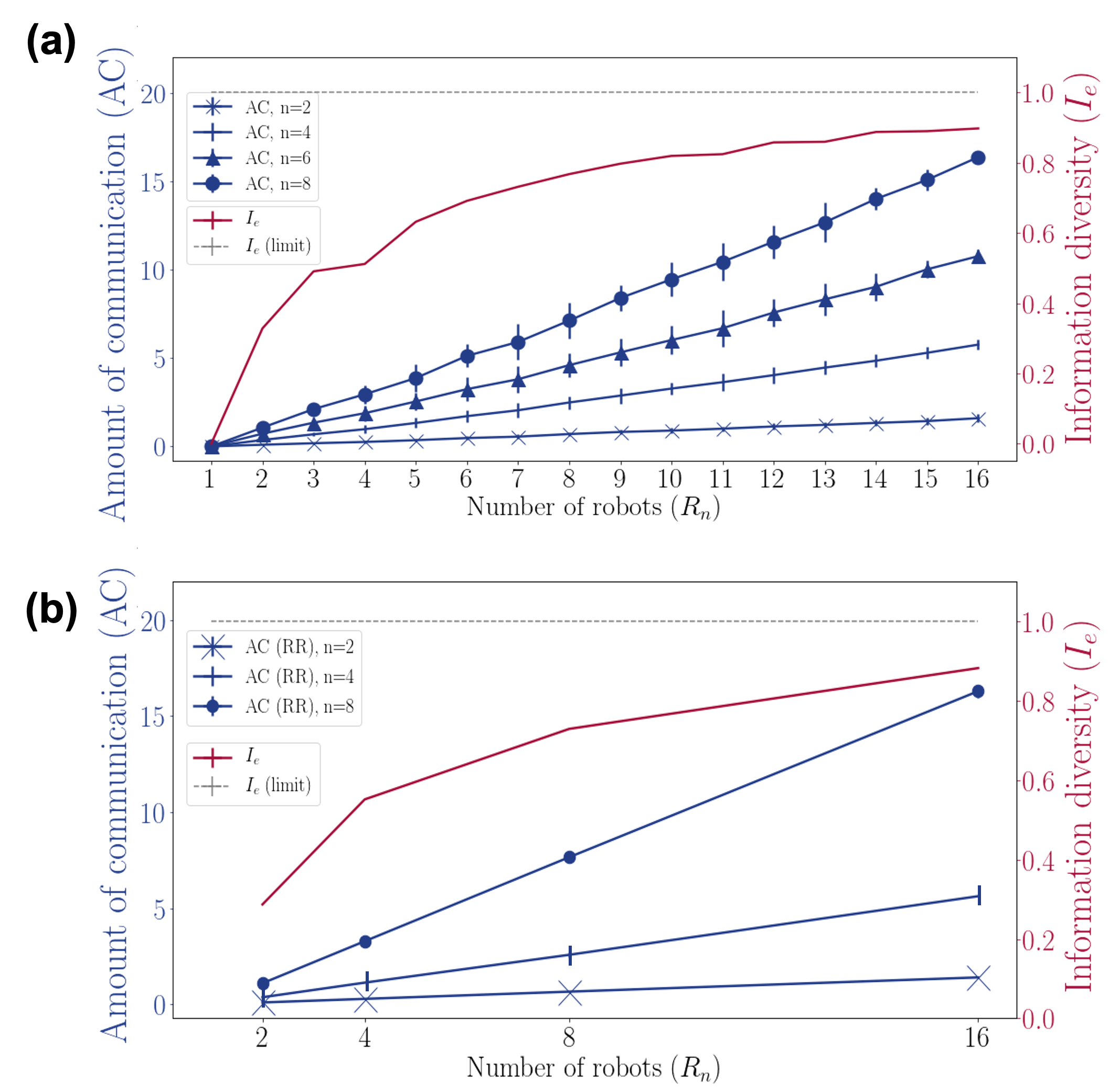}
\caption{\textbf{Amount of communication ($AC$) and information diversity ($I_e$) metrics for the foraging mission}. The $AC$ and $I_e$ metrics are depicted in blue and red colors, respectively, for different $R_{n}$ and $n$ values used in simulation (a) and real-robot (b) experiments. The upper-bound limit for the $I_e$ metric is depicted in both figures with a dashed grey line. Averaged results and standard deviations in both scenarios suggest a direct relationship between the two metrics in the foraging mission.}
\label{fig:CC_ID_Vs_NumberOfRobots}
\end{figure}

Fig.~\ref{fig:CC_ID_Vs_NumberOfRobots} shows average results and standard deviations for the amount of communication ($AC$) in KB and information diversity measured by Shannon's equitability index ($I_{e}$) for different simulation and real-robot experiments. In particular, Fig.~\ref{fig:CC_ID_Vs_NumberOfRobots}(a) reports simulation results for the $R_{n}$ ($1\leq R_{n}\leq 16$) and $n\in\{2,4,6,8\}$ configurations, while Fig.~\ref{fig:CC_ID_Vs_NumberOfRobots}(b) reports results of real robot experiments for $R_{n}=[2,4,8,16]$ and $n\in\{2,4,8\}$ configurations. In both figures, $I_{e}$ $\in$ [0, 1] gives information on the variability in the number of operations performed by each robot in the swarm during the mission. Lower values indicate more uneven distributions (e.g., one single robot completing all operations) while higher values indicate more uniform distributions (e.g., all robots completing the same number of operations). Figs.~\ref{fig:CC_ID_Vs_NumberOfRobots}(a-b) also show that $AC$ increases linearly with $R_{n}$ since there are more robots exchanging MT proofs. Moreover, MTs with larger $n$ values make the proofs robots exchange ``heavier'' (i.e., $\pi$ is longer), which also contributes to increase $AC$. However, larger $R_{n}$ values tend to increase information evenness ($I_e$) in the swarm. In other words, sequential operations are more evenly distributed. Results for both simulation and real-robot experiments suggests that, as we increase $R_{n}$, the number of operations fulfilled by the robots tends to become more uniform. Fig.~\ref{fig:CC_ID_Vs_NumberOfRobots} also suggests that information diversity might need a really large $R_{n}$ value to converge to 1 (i.e., to complete ``evenness''). 

\subsection*{Maze formation mission experiments}
\label{sec:Maze}

\begin{figure}[p]
\centering
\includegraphics[width=0.75\linewidth]{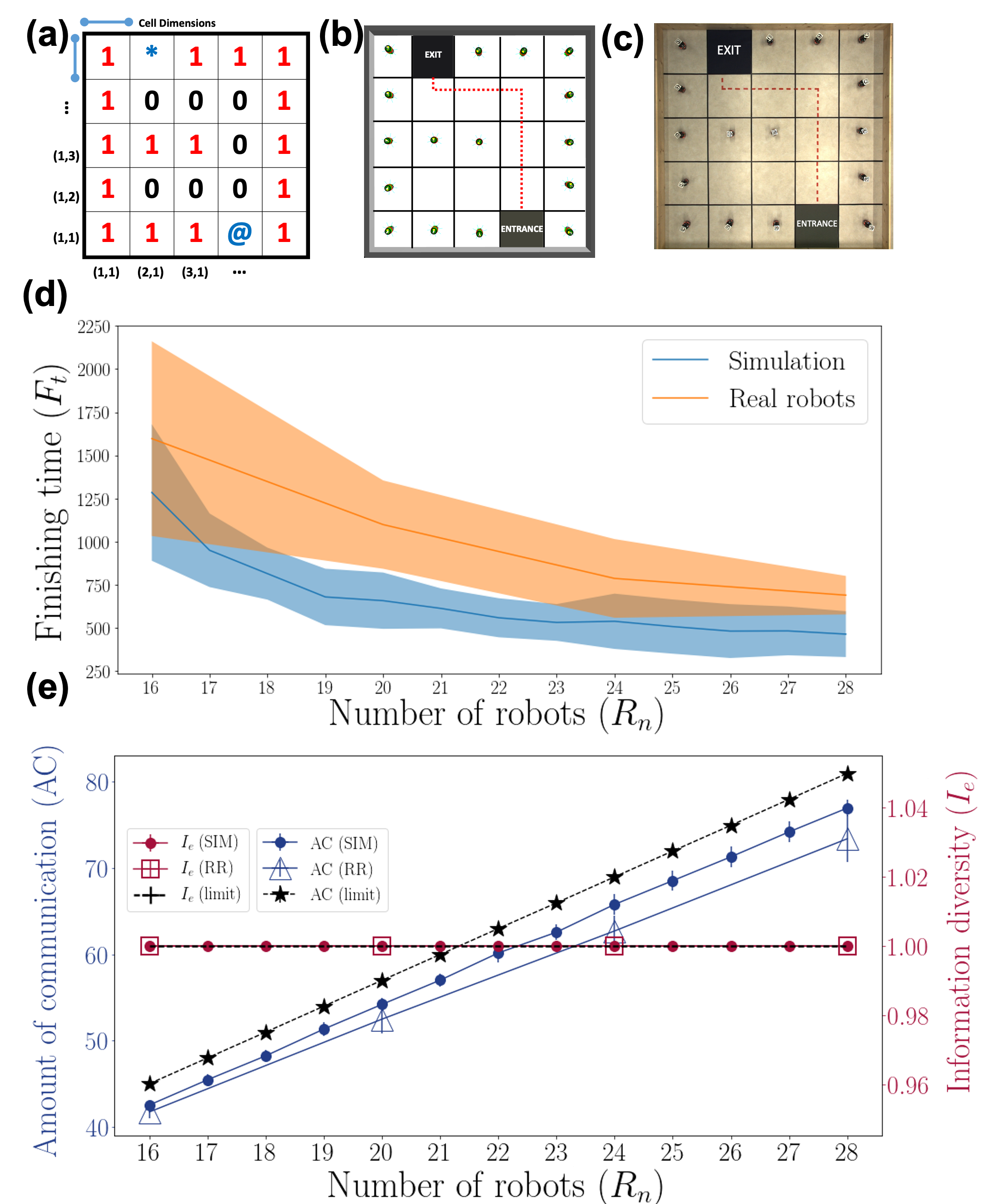}
\caption{\textbf{Maze formation scenario}. (a) $5\times 5$ matrix used to represent a maze. Four different elements are included in the array: 0 (black) represents an empty space, 1 (red) represents a wall, @ and * (blue) represent the entrance and the exit of the maze, respectively. (b) The maze depicted in (a) built by a swarm of simulated robots (a typical simulation run is displayed in \href{https://youtu.be/QT-TE4BqCyk}{movie M3}) where the path between entrance and exist is depicted with a dashed line. (c) The maze depicted in (b) built by a swarm of real e-puck robots (a typical real-robot run is displayed in \href{https://youtu.be/8Rq2wYWtNIk}{movie M4}). (d) Average $F_{t}$ (in seconds) and its standard deviation for the maze formation mission with different robot swarm sizes for both simulation (blue) and real-robot (orange) experiments. All results were obtained with a 16 operations ($n=16$) maze as shown in (a-c). (e) Amount of communication ($AC$) and information diversity ($I_e$) metrics in blue and red colors, respectively, for both simulation (circles) and real robots (triangles and squares) and for different $R_{n}$ values for the maze formation mission. Upper-bound limit values for both $AC$ (star) and $I_e$ (line) are depicted with dashed black lines respectively. Average results suggest no relationship between the two metrics.}
\label{fig:Maze}
\end{figure}

In the second experiment, we gave the robot swarm a maze formation mission (Fig.~\ref{fig:Maze}(a)-(c)) where we conducted 100 simulations and 15 real-robot experiments for each considered configuration. Fig.~\ref{fig:Maze}(a) represents the ``blueprint'' of a $5\times 5$ maze where 0 represents an empty space, 1 a wall, and * and @ the entrance and the exit of the maze, respectively. As in the foraging mission, robots first wander around the arena. However, instead of looking for colored cells, in this mission robots search for (x,y) maze coordinates. By knowing the cell dimensions ($0.5\times 0.5$ m$^{2}$) robots can calculate the (x,y) coordinates of cells they are located at. Every time a robot discovers a new (x,y) coordinate pair (e.g., (1,1)), it uses it to generate $h_{s}$. Then, it generates the value $h_{a}$ for all possible actions.\footnote{In the considered maze formation mission, $h_{a}$ is restricted to the ``stop'' action.} Like in the foraging mission, $H_{i}$ is calculated by hashing $h_{s}$ and $h_{a}$: $H_{i} = H((h_{s},h_{a}))$ and verified against the current $i$ of the robot. In case it is possible to generate a valid proof $\pi$ with this information ($\exists\pi_{(i,H_{i})}$) the robot finds the center of the cell and stops there (Fig.~\ref{fig:Maze}(b-c)). If not, the robot keeps wandering. In the same way as in the foraging mission, robots avoid already completed operations (in this case, they avoid stopping in already occupied cells) by receiving the proof that $i$ was already completed. Additional information about robot's behavior in the maze formation mission can be found in the Supplementary Materials.

As with the foraging experiment, the maze formation mission is finished once all operations have been completed. In this maze formation mission, $n$ was fixed to 16 in order to match the number of cells where the value 1 is present in Fig.~\ref{fig:Maze}(a). In addition, we explored a wider $R_{n}$ range (16 $\leq$ $R_{n}$ $\leq$ 28). Fig.~\ref{fig:Maze}(d) shows average $F_{t}$ and standard deviations for the maze formation mission for both simulation and real-robot experiments: as it was the case in the corresponding foraging experiment (Fig.~\ref{fig:Foraging}(a)), larger $R_{n}$ values reduce $F_{t}$. However, beyond a certain $R_{n}$ value ($R_{n}$ $\geq$ 24), the impact on $F_{t}$ becomes very small. Fig.~\ref{fig:Maze}(e) shows the $AC$ and $I_e$ metrics for the maze formation mission. Fig.~\ref{fig:Maze}(e) follows the same pattern as Fig.~\ref{fig:CC_ID_Vs_NumberOfRobots}: $AC$ increases linearly with $R_{n}$. Fig.~\ref{fig:Maze}(e) includes the $AC$ upper limit following equation~\ref{eq:CC} (see Statistical methods). According to Fig.~\ref{fig:Maze}(e), both simulation and real-robot $AC$ fall under the limit and suggest that real-robot experiments require a lower amount of communication than their simulation counterparts. Finally, Fig.~\ref{fig:Maze}~(e) shows that in this scenario complete ``evenness'' of information (i.e., $I_{e}=1$) is achieved. 

\subsection*{Experiments with larger-dimension missions}
\label{sec:LegoProjections}

\begin{figure}[tbh]
\centering
\includegraphics[width=\linewidth]{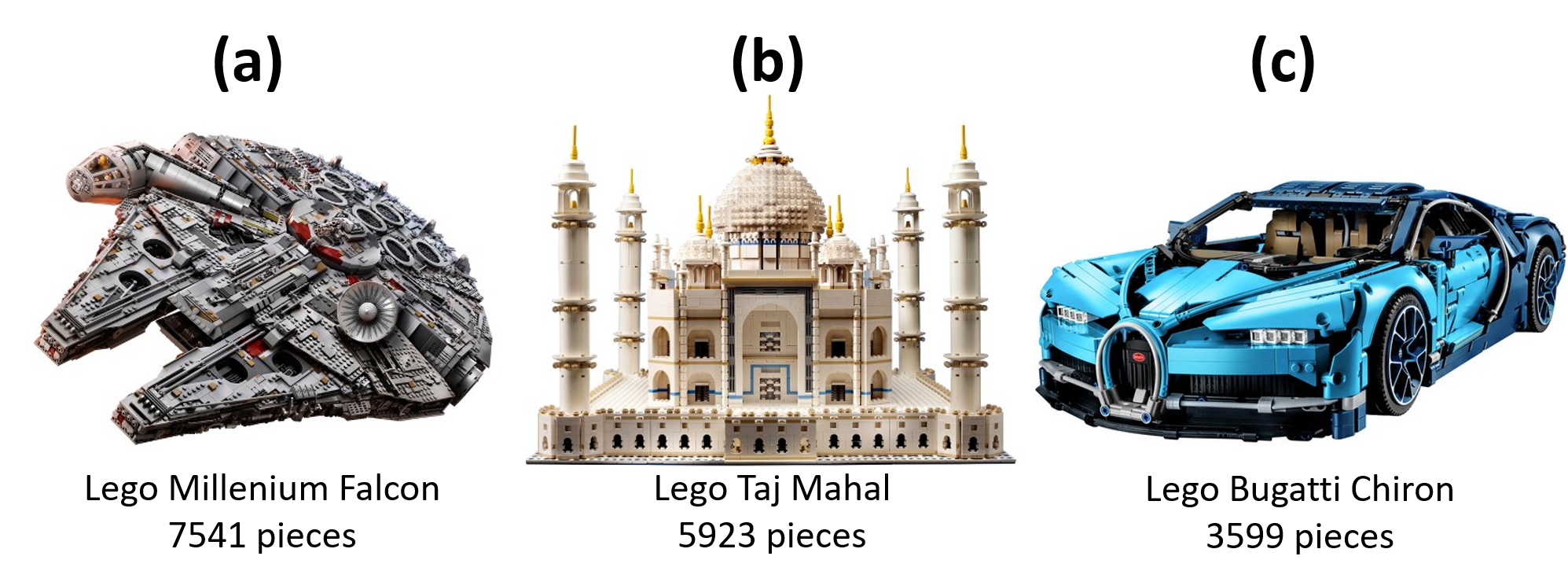}
\caption{\textbf{LEGO$^{\textregistered}$ models with their corresponding piece count}. LEGO$^{\textregistered}$ models are a good example of complex sequential missions that could be encapsulated in Merkle trees.}
\label{fig:Lego}
\end{figure}

\begin{table}[h!]
\begin{center} 
\caption{Memory, maximum amount of communication per robot ($AC/R_{n}$), and computation times for the models depicted in Fig.~\ref{fig:Lego}. Time measures (in seconds) are averages (standard deviations in brackets) over 100 runs using the Gumstix Overo board, the on-board computer mounted on the e-puck robots.}
\label{tb:LegoProjections}
\begin{tabular}{|m{0.25\linewidth}|m{0.15\linewidth}|m{0.15\linewidth}|m{0.25\linewidth}|}
\hline
\begin{center} {Model in Fig.~\ref{fig:Lego}} \end{center} & \begin{center} {Memory} {\footnotesize(KB)} \end{center} & \begin{center} { $AC/R_{n}$} {\footnotesize(MB)} \end{center} & \begin{center} { Average computation times$^*$} {\footnotesize(s)} \end{center} \\
\hline
{\footnotesize (a) Millennium Falcon} \newline {\footnotesize (7541 pieces)} & {\footnotesize 235 KB}  &  {\footnotesize 3.39 MB} & 
\shortstack{{\footnotesize \textbf{G:} 0.35 (0.001)}\\ {\footnotesize \textbf{P:} 0.0017 (0.0002)}\\ {\footnotesize \textbf{V:} 0.002 (7.96 \num{e-5})}}\\
\hline
{\footnotesize(b) Taj Mahal} \newline {\footnotesize(5923 pieces)} & {\footnotesize185 KB}  &  {\footnotesize2.54 MB} & \shortstack{{\footnotesize\textbf{G:} 0.28 (0.004)}\\ {\footnotesize\textbf{P:} 0.0016 (7.40 \num{e-05})}\\ {\footnotesize\textbf{V:} 0.002 (0.0003)}} \\
\hline
{\footnotesize (c) Bugatti Chiron} \newline {\footnotesize (3599 pieces)} & {\footnotesize 112 KB}  &  {\footnotesize 1.46 MB} &  \shortstack{{\footnotesize \textbf{G:} 0.16 (0.001)}\\ {\footnotesize \textbf{P:} 0.0015 (3.61 \num{e-05})}\\ {\footnotesize \textbf{V:} 0.0018 (9.37 \num{e-5})}}\\

\hline
\end{tabular}
\vspace{-2mm}
\begin{itemize}
   \scriptsize{\item[$^*$] \textbf{G}: time to generate the complete Merkle tree; \textbf{P}: time for the generation of a proof; \textbf{V}: time for the verification of a proof.} 
  \end{itemize}
\end{center}
\end{table}

Encouraged by these results, we found appropriate to analyze the feasibility of our approach in complex missions where the number of operations takes relatively large values. Fig.~\ref{fig:Lego} shows different LEGO$^{\textregistered}$ models where a sequential set of operations is required to achieve the final outcome (i.e., build the replica). These models\footnote{Three of the replicas with the largest piece count according to the current LEGO$^{\textregistered}$ catalog: \href{https://shop.lego.com/}{https://shop.lego.com/en-US/}} are good projections of the missions presented previously, especially since $n$ takes a relatively large value. Due to the possibility of accurately calculating the upper limit of $AC$ required to make all robots complete their MTs (see Statistical methods, Eq. \ref{eq:CC}) as well as the overall size of the MT stored by robots, we can compute Fig.~\ref{fig:Lego}'s corresponding MTs and measure their memory, maximum amount of communication per robot ($AC/R_{n}$) requirements. We also computed the average computation time to generate the complete Merkle tree (G), the time for the generation of a proof (P), and for the verification of a proof (V). Results for the aforementioned models are given in Table \ref{tb:LegoProjections}.

%The discussion describes the conclusions that can be drawn from the results, as well as the significance and implications of the research. A paragraph discussing the limitations of the study should be included and any issues that will need to be addressed before application to animal, human, or environmental health should also be described. 
\section*{Discussion}
\label{sec:Discussion}

We showed that two of the main MT properties (i.e., correctness and security) open a new path towards secure and secret cooperation in robot swarms. Regarding the security aspect, by using this approach, in order to cooperate, the robots in a swarm are required to prove to their peers that they fulfilled certain actions or that they know or own some particular information (i.e., proof-of-ownership \cite{Halevi2011}), rather than merely rely on information received from other robots (sensor data, votes, etc.). This approach makes robots resistant against potential threats such as tampering attacks since any change in the operation's data (i.e., $s$, $a$) will necessarily change the proof's outcome. Regarding the secrecy component, with the use of MTs robot swarms are now able to separate the mission data from its verification. This allows robots to verify that an operation was carried out by a member of the swarm without knowing what this operation entailed or which robot took part in its completion. This makes physical capture attacks inefficient since individual robots might not have enough raw or unprotected information to describe the high-level swarm's missions and goals, especially in large systems. However, this does not prevent robots in a swarm from cooperating to fulfill complex missions since robots can still prove to their peers that certain operations were identified and completed.

This approach was tested in two different scenarios: a foraging and a maze formation mission. In the foraging case, results suggest that increasing the swarm size has a positive impact on the performance of the system: larger swarm size have lower finishing time ($F_{t}$) and a higher probability of success ($P_{s}$). Results also show that the amount of communication ($AC$) grows linearly with the swarm size ($R_{n}$), which in extreme situations (e.g., very large swarms) could be detrimental to the system since individual robots might not be able to cope with the bandwidth requirements. In contrast, increasing $R_{n}$ has the positive effect of increasing the information diversity ($I_e$) since we are increasing the probability of reaching more uniform distributions of completed operations within the swarm. 

In the maze formation mission, where $R_{n}$ and $n$ take larger values, results also suggest that $R_{n}$ maintains an inverse relationship with $F_{t}$. In the maze formation mission, $I_e$ is maximized (i.e., $I_e$ = 1), which is possible since in this mission, when robots find a cell where they can generate a valid $\pi$ proof, they stop at its center, in this way making robots capable of fulfilling only one operation per mission, in contrast to the foraging scenario, where one robot might be able to complete several operations. Even though $AC$ grows linearly with $R_{n}$, this still does not represent a challenging situation for the swarm (e.g., 72 and 70 KB are exchanged in a 28 robots system for both simulation and real-robot experiments). Counterintuitively, real-robot experiments require a lower amount of $AC$ to complete the maze-formation mission than their simulation counterparts. This effect is due to the incapability of the simulation engine to take into account network lag: delays in the network produce situations where robots are able to find and complete the mission, while other robots in the swarm are still trying to synchronize on previous completed operations. Finally, it is interesting to emphasize that, due to these properties, robot swarms can complete complex missions such as the maze formation one without the means to infer high-level details such as where the entrance or the exit might be located.

Finally, Table \ref{tb:LegoProjections} shows that neither the memory, nor the communication requirements per robot, nor the average computation times of the corresponding MTs are out of reach of current commodity hardware (e.g., Overo Gumstix) and therefore our approach is feasible for current robot platforms. It is important to note that more than 99\% of the computation time is taken by the generation (G) of the MT that only takes place at the beginning of the mission, while the proof assembly (P) and validation (V) take a nearly insignificant amount of time. 

\newpage

\subsection*{Limitations and future work}
\label{sec:Limitations}

% Limitations and possible improvements 
\textbf{\normalsize The complete set of operations (and its order) must be known at design time} 

It is important to note that, as described previously, the current approach needs a swarm operator to design and encode the complete sequence of operations included in the MT before the mission starts. Therefore, any deviation from the original sequence included in the MT might be difficult to adapt to. \textbf{\small Possible improvement:} it is possible to make incremental updates to the mission and encapsulate the outdated MT within a newer version that takes into account all necessary changes. In that situation, a new MT root might be necessary to reflect the updated mission sequence. However, the most critical point for this solution is to trust the source that distributes the new root node hash. Fortunately, new methods such as blockchain-based communication are showing a promising way to make robots rely on trusted sources \cite{Strobel2018, Strobel2020}. 

\noindent\textbf{\normalsize Generation of a valid MT proof does not imply execution}

Another limitation of the system is that even if a robot discovers a valid combination of $h_{a}$ and $h_{s}$ values, thus being able to generate a valid MT proof, there is no guarantee the robot fulfills the corresponding operation in the physical world (e.g., deliver discovered colors to a target location or stop at the right coordinates). \textbf{\small Possible improvement:} it is possible to add additional variables to the operation encoding process, for instance, forcing the robot to do the right action $a$, with the right sensor input $s$, at the right location $l$ such as $H_{i} = H((h_{s},h_{a},h_{l}))$. Another possible way to tackle this problem is the addition of robot ``validators'', which verify that every operation claimed by the robots is completed. 

Perhaps the most promising direction for future research is the possibility to implement missions where heterogeneous swarms with different sensing, computation, and actuation capabilities can cooperate and collaborate. Along these lines, the increasing amount of attention that swarm robotics is gaining envisages a future where different swarms run by different types of institutions (e.g., private, public) can co-exist in a same location (e.g., urban and disaster areas, battlefield). The method described in this research offers a path to make different swarms cooperate without exposing sensitive data about their internal processes, goals, and organizations. Finally, a new type of service where the robot commands are encapsulated within an MT and whose output is the cryptographic proof that the job has been completed is now a possibility. In the last section of the supplementary materials of this work is described an early-stage example in this direction where a web-based marketplace allows a robot swarm to form shapes on-demand. In the example, potential customers upload an MT describing the desired shape and send it to a smart-contract in the Ethereum blockchain, which returns a proof when the work is being completed by the robots.

% MATERIALS AND METHODS
% The materials and methods section should provide sufficient information to allow replication of the results. Begin with a section describing the objectives and design of the study as well as pre-specified components.

% In addition, include a section (when appropriate) that fully describes the statistical methods with enough detail to enable a knowledgeable reader with access to the original data to verify the results. The values for N, P, and the specific statistical test performed for each experiment should be included in the appropriate figure legend or main text.

\section*{Materials and Methods}
\label{sec:Materials and Methods}

\subsection*{Merkle trees for robot swarms}
\label{sec:MerkleTrees}

A Merkle tree (MT) \cite{Merkle1988} is a hash-based tree structure where data is not stored in the inner nodes but in the leaves (Fig.~\ref{fig:GeneralOverview}(b)). Every leaf node (i.e., operation) encapsulates the combined hash of two hashes: $h_{s}$ (hash of the robot sensor's input) and $h_{a}$ (hash of the robot's action). These two hashes describe an atomic step within the swarm's high-level mission. For instance, the hash of the action ``push'' ($h_{a}$) and the hash of the sensor input ``block (1)'' ($h_{s}$) are included in the first operation by using the hash function $H$: $H(h_{a}, h_{s})$. In this work, we assume that an external entity (e.g., the swarm's operator) encodes a valid MT before the swarm robotics mission can take place. This MT contains all needed information about the operations that the swarm has to execute to fulfill the mission in the correct order (Fig.~\ref{fig:GeneralOverview}(c) - I). The resultant MT is broadcast to all the robots (Fig.~\ref{fig:GeneralOverview}(c) - II) before the mission starts (Fig.~\ref{fig:GeneralOverview}(c) - III). Since MTs are an encrypted data structure, robots do not have direct access to the raw information concerning the mission, but only to the hash values comprising the MT are known.

At the beginning of the mission, all robots aim to fulfill the first operation. However, only one will succeed in doing so. The successful robot, first, wanders around the environment; second, when it perceives a new sensor input (e.g., block (1)) it generates a hash ($h_{s}$) from it; third, it generates another hash ($h_{a}$) for an action it can perform (e.g., ``push''); fourth, it combines $h_{s}$ and $h_{a}$ and rehashes them to obtain the hash string $H_{(h_{s},h_{a})}$; fifth, it matches $H_{(h_{s},h_{a})}$ against the hash value that corresponds to the first operation of the MT. At this point, the robot executes the resulting command (e.g., ``push block (1)''). The robot that fulfills this operation, can regard the operation as completed. Once the operation is completed, the (\emph{prover}) robot has the ability to generate cryptographic ``proofs'' ($\pi$) that demonstrate to other (\emph{verifier}) robots that it actually carried it out that operation because it has information corresponding to that operation that it could not have otherwise (i.e., $h_{s}$ and $h_{a}$). In this case, the proof $\pi$ is a list of hashes that \emph{verifier} robots can use to calculate the MT root node hash (shared by all robots in the swarm) by using $h_{a}$ and $h_{s}$ as inputs. These proofs can be sent and received by robots in the swarm, allowing them to synchronize, check, update, and complete their corresponding MT until the whole mission is completed. It is important to note that these proofs can be exchanged without disclosing any raw or unprotected data about the mission itself. The typical workflow between \emph{prover} (P) and \emph{verifier} (V) follows the pattern depicted in Fig.~\ref{fig:GeneralOverview}(d): (1) V sends a query Q to P regarding a particular operation in the MT (e.g., operation 0). (2) P sends a proof $\pi$ (list of hashes) that demonstrates knowledge about the requested operation. (3) V verifies the proof by computing in a bottom-up fashion the received information. The proof is regarded as valid if V can generate the root node hash it keeps in its memory (shared by P and V) by using $\pi$. More details about MT structure, proofs, and data encapsulation are provided in the Supplementary Materials.

\newpage

\subsection*{Statistical methods}
\label{sec:AnalysisMetrics}

To evaluate and analyze our approach we rely on three metrics: 

\textbf{Performance}. These measures show the speed and reliability with which a mission is carried out. In this paper, we use the mission's finishing time $F_{t}$ to measure the amount of time required to fulfill the swarm's mission. In addition, we use an estimate $P_{s}$ of the probability that the system attains its target objective in an amount of time $\tau$ \cite{Garattoni2018}. Formally, let $j \in \{1, \dots, k\}$ be the index of an experiment, $r_{j}$ be the run time of experiment $j$, and experiments where $r_{j} < TC$ are considered successful experiments. The estimate $P_{s}$ of the probability of success of the system over time (up to TC) is defined as $P_{s}(\tau\leq t)=\{j | r_{j}\leq t\}/k$.

\textbf{Amount of communication}. The amount of communication ($AC$) is computed by multiplying the number of times the \emph{prover}-\emph{verifier} workflow (Fig.~\ref{fig:GeneralOverview}(d)) takes place during the mission by the size (in bytes) of the proof ($\pi$) the robots exchanged. When the MT is perfectly balanced (i.e., the number $m$ of leaves is a power of 2), the size of the proof $\pi$ is $log_{2}(m)+2$: the number of hashes to reach the root node plus the $h_{s}$, $h_{a}$ hashes. In missions with $n$ operations where all robots need to complete their MTs, an upper limit  $AC_{ul}$ of the $AC$ metric can be calculated with the following equation: 

\begin{equation}  
AC_{ul} = P_{n} \cdot P_{l} \cdot \rvert H \lvert
\label{eq:CC}
\end{equation}

\noindent where $P_{n}=((R_{n}-1)\cdot n)$ is the total number of proofs exchanged, $P_{l}=log_{2}(\hat{n})+2$ is the length of the proof, $\hat{n}$ is the smallest power of 2 that is greater or equal to $n$, and $\rvert$H$\lvert$ is the size (in bytes) of the hash function used. The hash function used in this work (SHA256) has a hash size ($\lvert$H$\rvert$) of 32 bytes. It is important to note that the hash size can be increased (e.g., SHA3-512, SHA3-1024) in order to achieve improved security.

\textbf{Information diversity}. In this research, robots are only in contact with the raw sensor and action information from the operations they carry out themselves (i.e., input data for the $h_{a}$ and $h_{s}$ hashes). However, in swarm robotics applications, it is difficult to fully ensure a uniform distribution of operations or tasks among the robots. In this scenario, certain robots might be able to accumulate raw or unprotected information that could be stolen if they are subject to attacks (e.g., physical capture). To analyze this phenomenon, we introduce Shannon's equitability ($I_{e}$) to measure ``evenness'', that is, to measure how widely spread raw information is within the swarm. 

Shannon's equitability ($I_{e}$) can be calculated by dividing Shannon's index $I$ by $I_{max}$: 

\begin{equation}
I_{e} = \frac{I}{I_{max}}
\end{equation}

\noindent where Shannon's index $I = - \sum_{i=1}^{S} p_{i} \ln p_{i}$ (i.e., Shannon's entropy \cite{Shannon1948}) is a mathematical measurement used to characterize diversity in a community. In this case, $S$ represents the total number of robots that took part in the mission ($R_{n}$), $p_{i}$ is the percentage of operations robot $i$ conducted compared to the mission's length ($n$), and $I_{max} = \ln S$. $I_{e}$ assumes a value between 0 and 1. Lower values indicate more uneven distributions while higher values indicate more uniform distributions. In this case, 1 represents complete ``evenness'': all robots carried out the same number of operations and therefore were exposed to the same amount of raw information.

\nocite{Pinciroli2012}
\bibliographystyle{Science}

\newpage

\vspace*{-20mm}

\section*{Acknowledgment}
\label{sec:acknowledgment}
\textbf{Funding:} This project has received funding from the European Union’s Horizon 2020 research and innovation programme under the Marie Skłodowska-Curie grant agreement No. 751615. Marco Dorigo acknowledges support from the Belgian F.R.S.-FNRS, of which he is a Research Director. \textbf{Author contributions:} E.C.F. designed and realized the system as well as performed the experiments. E.C.F. and M.D. analyzed the results and wrote the manuscript. T.H., A.P., and M.D. directed the project. \textbf{Competing interests:} The authors declare that they have no competing interests. 

%Here you should list the contents of your Supplementary Materials -- below is an example. 
%You should include a list of Supplementary figures, Tables, and any references that appear only in the SM. Note that the reference numbering continues from the main text to the SM.
% In the example below, Refs. 4-10 were cited only in the SM.     

\section*{Supplementary materials}
Section S1. Detailed description of an MT structure, proof, and data encapsulation\\
Section S2. Detailed description of the real-robot experimental setup\\
Section S3. Detailed description of the simulation experimental setup\\
Section S4. Description of the finite state machines controlling the robots\\
Section S5. IRIDIA Swarm Marketplace\\
Figure S1. Description of a typical MT structure and proof\\
Figure S2. Robot interaction and FSM diagrams\\
Figure S3. Experimental setup for the real-robot experiments\\
Figure S4. IRIDIA Swarm Marketplace\\
Movie M1. Typical simulation run for the foraging experiments\\
Movie M2. Typical real-robot run for the foraging experiments\\
Movie M3. Typical simulation run for the maze formation experiments\\
Movie M4. Typical real-robot run for the maze formation experiments\\
Movie M5. Tangible Swarm (Supporting features)\\

% The following parameters seem to provide a reasonable page setup.
\topmargin 0.0cm
\oddsidemargin 0.2cm
\textwidth 16cm 
\textheight 21cm
\footskip 1.0cm

% Reset countering
\setcounter{figure}{0}   
\setcounter{page}{1}   
\setcounter{section}{0}   
\setcounter{table}{0}   

% Change numbering for SM
\renewcommand{\thepage}{S\arabic{page}} 
\renewcommand{\thesection}{S\arabic{section}}  
\renewcommand{\thetable}{S\arabic{table}}  
\renewcommand{\thefigure}{S\arabic{figure}}

%%%%%%%%%%%%%%%%% END OF PREAMBLE %%%%%%%%%%%%%%%%
%\begin{document} 

% Double-space the manuscript
\baselineskip 24pt

\section*{Supplementary materials}

\subsection*{S1 - Detailed description of an MT structure, proof, and data encapsulation}
\label{sec:MerkleTrees}

\begin{figure}[tbh]
\begin{center}
\includegraphics[width=\textwidth]{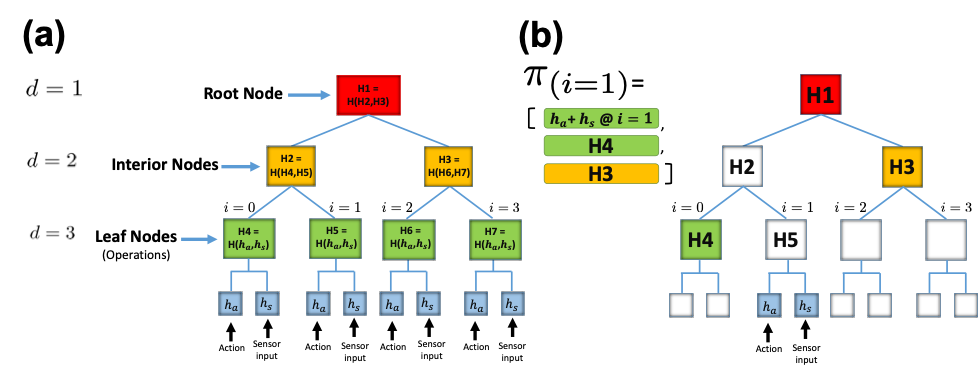}
\caption{\textbf{Description of a typical Merkle tree (MT) structure and proof}. (a) Merkle tree implementation ($d=3$) with 4 leafs, 2 interior, and 1 root nodes. Each leaf node (green) encapsulates the hash of two hashes: a robot action ($h_{a}$) and a robot sensor input ($h_{s}$). Each leaf node represents one atomic operation that robots should complete sequentially ($i=0$, $i=1$, $\dots$, $i=m-1$) in order to fulfill the swarm's mission. Interior nodes encapsulate the hash of their two children, and the root node encapsulates the hashes of its two interior nodes. (b) Describes the path (in color) in order to get the proof for leaf node $i=1$.}
\label{fig:MerkleTreeExplanation}
\end{center}
\end{figure}

A Merkle tree (MT) of depth $d$ is a tree with $m=2^{d-1}$ leaf nodes, each one with an index $i$ $\in [0, m-1]$. Each leaf node encapsulates a hash string of an associated operation, while each interior node contains the hash of the combination of its two children. A depiction of a complete MT for $d=3$ is given in Fig.~\ref{fig:MerkleTreeExplanation}(a). Each operation (green blocks) encapsulates the combined hash of two hashes: $h_{s}$ (hash of the sensor's input $s$) and $h_{a}$ (hash of the robot action $a$). These two hashes describe an atomic step within the swarm's high-level mission. For instance, the hash of the action ``push'' ($h_{a}$) and the hash of the sensor input ``block (1)'' ($h_{s}$) would be included in one of the operations by using the hashing function $H$: $H(h_{a}, h_{s})$. As outlined in Fig.~\ref{fig:GeneralOverview}(d), when a \emph{verifier} ($V$) robot queries a \emph{prover} ($P$) robot with updated information (i.e., $P$'s index $i$ is greater than $V$'s index $i$), $P$ returns a chain of digests $\pi$ needed to compute the root node digest (red block). $V$ keeps (at least) a copy of the root node hash itself (computed before the experiment starts), and checks $\pi$ by trying to recompute the root node hash in a bottom-up manner. 

Fig.~\ref{fig:MerkleTreeExplanation}(b) shows the proof $\pi$ for a fetch at operation $i=1$. It consists of four elements in sequence, the two hashes $h_{a}$ and $h_{s}$, the hash H4, and the hash H3. Then, the verification proceeds bottom up: $V$ computes the hashes of the two hashes $h_{s}$ and $h_{a}$, which is H5, and concatenates it with the hash of H4 provided in $\pi$. Next, it concatenates H4 and H5 and computes the hash of the digest for node H2. Then, it concatenates H3 provided in $\pi$, with the computed H2, and hashes the result. Finally, it checks whether this computed digest equals H1 (root node hash). 

In this research, we are interested in two properties derived from the workflow between $P$ and $V$ robots depicted in Fig.~\ref{fig:GeneralOverview}(d): correctness and security. First, correctness implies that when $P$ executes a query $Q$ over its own MT, the generated proof $\pi$ requires knowledge of $h_{a}$ and $h_{s}$ and can be easily verified by $V$ locally. This property opens the path to secret cooperation between robots since encrypted verification and validation of data (e.g., by using cryptographic hashes) can be exchanged within the swarm without disclosing any raw or unprotected information. Second, security implies that a computationally limited, deceiving $P\prime$ cannot induce $V$ to accept a faulty answer. The basis of this property is the use of collision-resistant hashes: if $P\prime$ can cause $V$ to accept an incorrect answer then the proof returned by $P\prime$ will yield a collision (i.e., two different inputs produce the same output hash value). This research assumes the hash generation method used (SHA256) is collision-free, that is, that the probability that different inputs produce the same output is negligible. This property leads to secure cooperation between robots.

\subsection*{S2 - Detailed description of the real-robot experimental setup}
\label{sec:RealRobotArenaSetup}

\begin{figure}[tbh]
\centering
\includegraphics[width=0.95\linewidth]{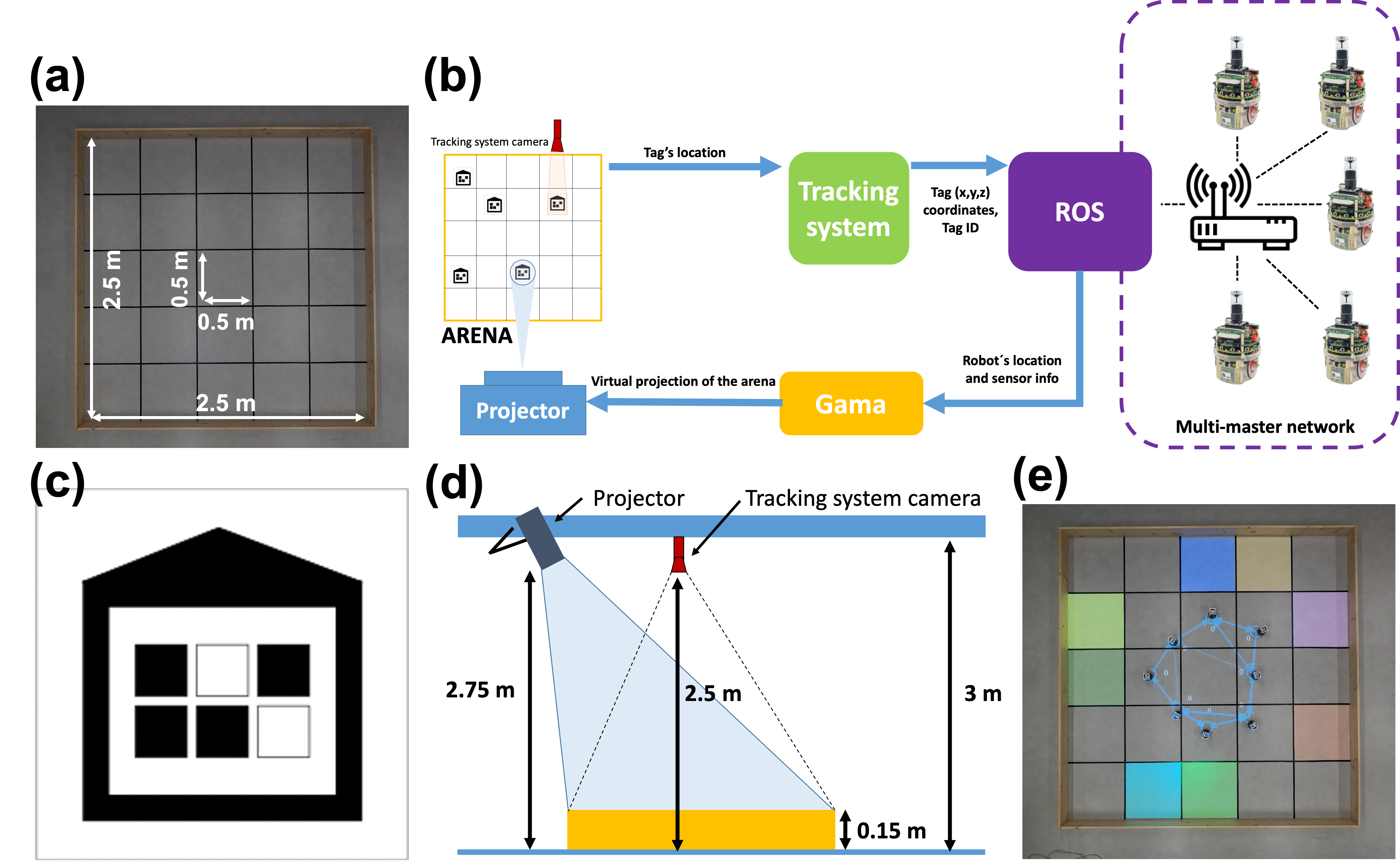}
\caption{\textbf{Experimental setup for the real-robot experiments.} (a) arena (with measures) used during the real-robot experiments. (b) Information workflow for the real-robot experiments. The diagram shows how robot tags' location are captured by a vision-based tracking system, coordinates for each tag are calculated, and used by the robots connected in a network. Then, relevant information such as robot's sensor input or status can be projected on the arena where experiments are being run in real-time. (c) Tag with ID number 34 encoded in binary as a matrix of black and white squares. (d) Diagram with the physical disposition of the sensing (camera) and actuation (projector) elements introduced in (b). (e) Image of a real-robot experiment for the foraging mission using the information workflow described in (b).}
\label{fig:ExperimentalSetupRealRobot}
\end{figure}

\noindent Fig.~\ref{fig:ExperimentalSetupRealRobot}(a) shows a bird's eye view of the arena built for the real-robot experiments. The arena is a rectangular area of 2.5 $\times$ 2.5 m$^{2}$ built by 4 wooden planks attached to each other using hot melt glue. Each one of these planks has the following measures: 2.5 m (length) $\times$ 0.15 m (height) $\times$ 0.05 m (depth). The experimental area is divided into a 5 $\times$ 5 grid where each cell corresponds to a 0.5 $\times$ 0.5 m$^{2}$ area. The black lines between the cells were created by using black duct tape which also covers the rim of the linoleum tiles underneath. 

Fig.~\ref{fig:ExperimentalSetupRealRobot}(b) shows the information workflow that takes place during the real-robot experiments. First, each robot carries a tag (Fig.~\ref{fig:ExperimentalSetupRealRobot}(c)) on top of its omnidirectional camera cylinder; this tag encodes a specific robot ID used in the system. Second, tags are recognized by a vision-based tracking system \cite{StrTurSalGarFraReiDorBir2013:techreport-013} using a camera (Prosilica GC 1600C) right on top of the arena (Fig.~\ref{fig:ExperimentalSetupRealRobot}(d)). Third, (x,y,z) coordinates for each tag are calculated with respect to the bottom-left corner of the arena and sent to a ROS \cite{Quigley2009} multi-master \cite{Tiderko2016} instance connected to the robots through a WiFi router. With this configuration, robots can access their own position and orientation and use them during the mission (e.g., locate the target location, find the center of each cell, etc.). The WiFi network is used as a low-level communication medium for the robots. In order to keep the decentralized aspect of the robot swarm, we designed a higher-level protocol where each robot uses its own ROS multi-master instance to communicate with other robots (e.g., send queries, receive proofs, etc.). This topology avoids that the system relies on a single ROS master for data exchange. Robots communicate in a peer-to-peer fashion with each other since their own multi-master instances can only exchange data if robots are within communication range ($C_{range}$). 

\subsubsection*{S2.1 - Supporting features of the system}
\label{sec:SupportingFeatures}

In order to better understand the robot swarm behavior during an experiment, we built a system capable of collecting relevant information---such as the robots’ internal states and the interaction patterns within the swarm---from the robots during the mission and of displaying it on the robot swarm working space in real-time. Following the workflow depicted in Fig.~\ref{fig:ExperimentalSetupRealRobot}(b), information about the robots' location, sensor input and internal state is extracted from the multi-master network at a frequency of 10 Hz and sent to an agent-based simulation software named Gama \cite{Drogoul2013:Gama}. With this information, Gama is able to create a virtual representation of the current state of the arena which is projected on the arena itself by using a short throw projector (Optoma EH460ST). Fig.~\ref{fig:ExperimentalSetupRealRobot}(d) shows a diagram of the arrangement of the tracking system camera and of the projector with respect to the arena. Fig.~\ref{fig:ExperimentalSetupRealRobot}(e) shows a real-robot foraging experiment where the virtual representation of the arena (e.g., colored cells, communication patterns between the robots, robot's internal variables, etc.) are projected on the arena and robots in real-time. Detailed information about the amount of information displayed in both foraging and maze-formation missions is included in the \href{https://youtu.be/hImae_ykp8M}{M2} and \href{https://youtu.be/8Rq2wYWtNIk}{M4} movies. 

The idea of building a close-loop control where a camera (e.g., performing tag tracking, pattern recognition, etc.) acts as the sensing part and a projector is the actuation component can be extended to other uses and scenarios. For that reason, we designed a tool named {\it Tangible Swarm}\footnote{\href{https://github.com/edcafenet/TangibleSwarm}{https://github.com/edcafenet/TangibleSwarm}} that has the aim to display in a real-time fashion relevant information about a swarm robotics system at the same place where the swarm is conducting its mission. Information such as IDs, sensor inputs, robot trajectories, distance between robots in a formation scenario, battery status, communication patterns, etc.~can be easily coded, displayed and customized for different robot swarm missions. \href{https://youtu.be/pUGIoen47jE}{Movie M5} shows several of the aforementioned capabilities and gives an overview of the tool capabilities. 

\subsection*{S3 - Detailed description of the simulation experimental setup}
\label{sec:ExperimentalSetup}

\begin{figure}[tbh]
\centering
\includegraphics[width=0.6\linewidth]{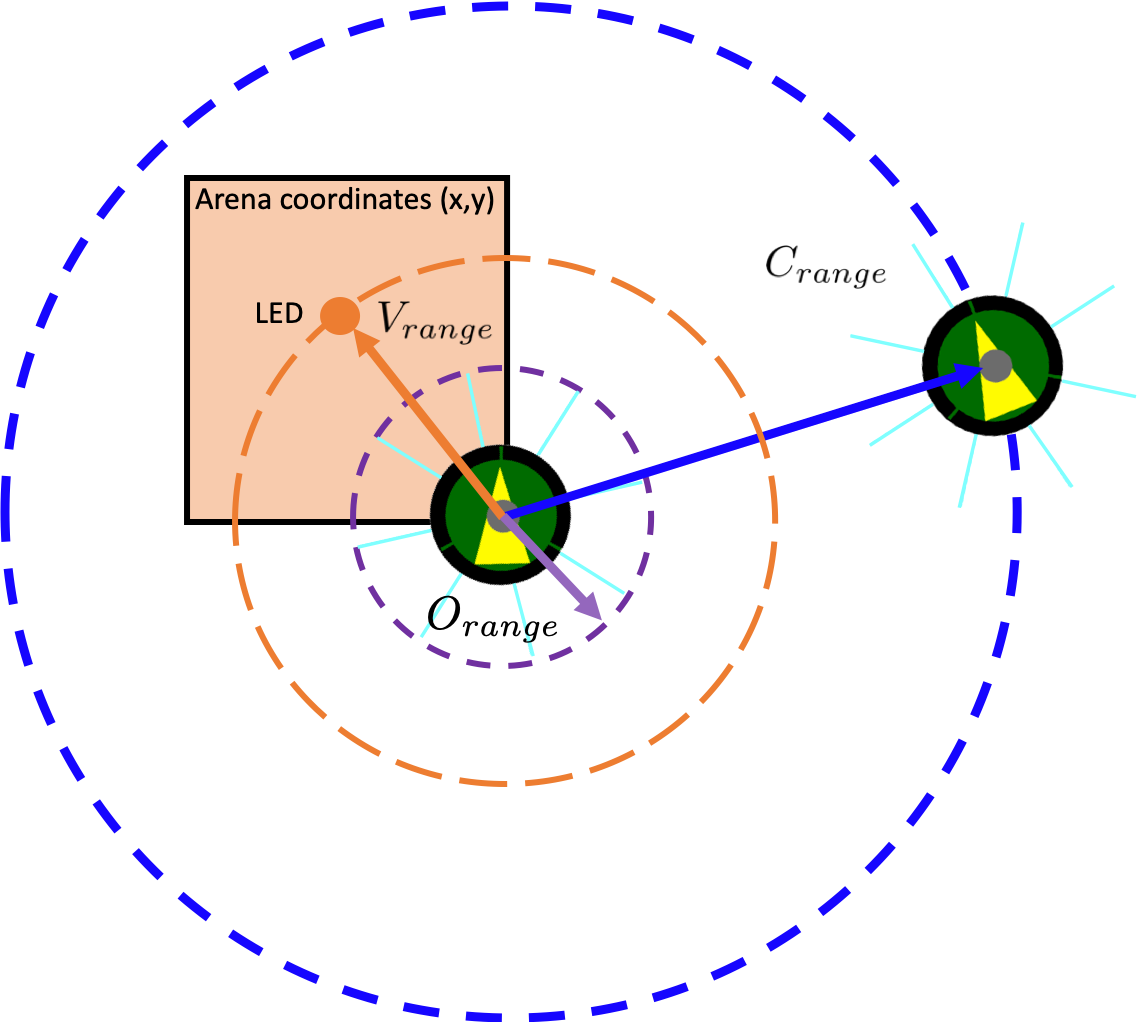}
\caption{\textbf{Robot interaction diagram.} Every robot controller relies on three main thresholds: $O_{range}$ represents the range where obstacles are detected by robots in order to avoid collisions. $V_{range}$ represents the robot's maximum vision range to detect tasks (i.e., see LED markers) in the arena. $C_{range}$ represents the maximum required distance for robots to communicate with one another. Finally, robots are able to locate themselves in the arena's 2D space either to find the target location (foraging mission) or to obtain the (x,y) coordinates of the grid (maze formation mission).}
\label{fig:InteractionDiagram}
\end{figure}

% Actual setup of the experiment
For the results obtained in the simulated foraging mission, the following parameters were used: the color $k$ of a task is chosen in the set $\{$green, magenta, blue, yellow, red, cyan, lime, orange$\}$. Only one task of each color $k$ is placed in the arena at the beginning of each run. For example, in the run where the MT has 4 operations ($n=4$), a total of 4 tasks were randomly scattered in the arena. For the robot communication and interaction space (as depicted in Fig.~\ref{fig:InteractionDiagram}) we used the following parameters: The robot communication range ($C_{range}$) was set to 1 m (maximum range of the range and bearing device of the e-puck robots), the vision sensing distance ($V_{range}$) was set to 0.35 m (maximum range to detect an LED marker from within the same cell). LEDs are used in the foraging mission to help robots visually locate the cell and define its color. The robot obstacle detection range ($O_{range}$) was set to 0.10 m. 
In addition, robots are able to locate themselves (Fig.~\ref{fig:InteractionDiagram}) in the 2D arena by using the simulator positioning sensor, this information is used to travel to the center of the cell of the target location. 
Finally, the Time Cap (TC) for each experiment was set to 5{,}100 seconds (maximum battery duration\footnote{According to manufacturer: \href{https://www.gctronic.com/doc/index.php/Overo\_Extension\#Consumption}{https://www.gctronic.com/doc/index.php/Overo\_Extension\#Consumption}}). All the parameters mentioned previously were used also for the maze formation mission, excluding those concerning the discovery of tasks. In contrast to the foraging mission, in the maze formation mission, robots are able not only to locate themselves in the 2D arena ($2.5\times 2.5$ m$^{2}$), but also to calculate the (x,y) coordinates of the grid which they used as input to generate the $h_{s}$ hash. All simulation experiments presented in this paper were conducted in ARGoS \textit{(38)}, a modular multi-robot simulation and development environment. 

\subsection*{S4 - Description of the finite state machines controlling the robots}
\label{sec:FSM}

\begin{figure}[tbh]
\centering
\includegraphics[width=0.6\linewidth]{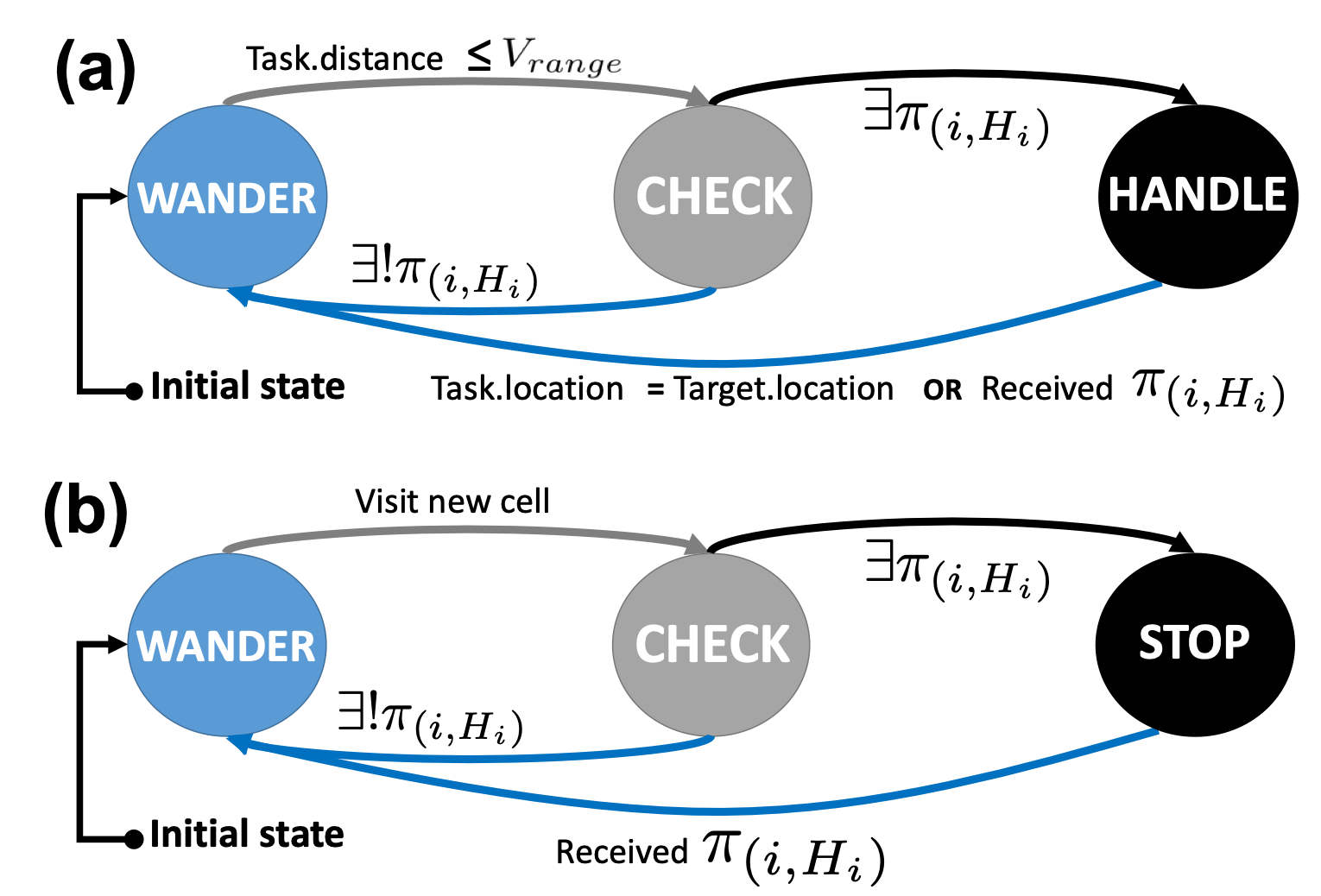}
\caption{\textbf{Robot FSM diagrams.} (a) Finite State Machine (FSM) controller for the foraging mission. Three behaviors ---\textbf{Wander}, \textbf{Check}, and \textbf{Handle}---are coded in every robot. (b) FSM controller for the maze formation mission. This FSM differs from the one in (a) in two ways. First, the \textbf{Handle} behavior is substituted by the \textbf{Stop} behavior, which makes robots stop at the center of a grid cell instead of carrying a token to the target. Second, the return condition to the \textbf{Wander} behavior is only executed if robots receive proof that the cell they are trying to occupy has been already filled.}
\label{fig:ExperimentalSetup}
\end{figure}

Fig.~\ref{fig:ExperimentalSetup}(a) depicts the finite state machine (FSM) that controls the robot for the foraging mission, which relies on three basic behaviors:

\textbf{Wander}. The robot performs a random walk searching for tasks in the arena. If the robot detects a task within its vision range (\texttt{Task.distance} $\leq V_{range}$), it executes the \textbf{Check} behavior; otherwise it continues searching. During the execution of this action the robot is able to detect obstacles such as walls or other robots (within distance $O_{range}$) and avoid them.
    
\textbf{Check}. Once a task is within the robot's vision range, the robot can sense the color of the LED marker associated with it. This information is used by the robots in order to generate the sensor input hash ($h_{s}$). For the considered foraging mission, the action hash $h_{a}$ encodes the ``carry to target'' action. Robots executing the \textbf{Check} behavior combine $h_{s}$ and $h_{a}$ to generate a hash ($H_{i} = H((h_{s},h_{a}))$) that is used to generate the proof $\pi$. In case a proof $\pi$ exists for the current working operation $i$: $\exists\pi_{(i,H_{i})}$, the combination of task color and action (i.e., operation) can be verified as part of the MT. Otherwise ($\exists!\pi_{(i,H_{i})}$), the robot returns to the \textbf{Wander} behavior. 

\textbf{Handle}. In case the robot generates a valid proof for the visible task, the robot travels to the center of the cell. Then, the robot delivers the color to the target located at center of the arena. Once the robot reaches its destination (\texttt{Task.location = Target.location}), the robot releases the task information and changes the status of $i$ as completed in its local MT. Then, the robot increments the pointer of the current working operation index: $i=i+1$ for $i$ $\in$ [0, n-1]. In case the robot receives a proof that the $i$ has been already completed by another robot while carrying the task, the robot updates its own MT with the received information and returns to the \textbf{Wander} behavior. 

Fig.~\ref{fig:ExperimentalSetup}(b) depicts the FSM that controls the robot for the maze formation mission. There are three differences with respect to the foraging mission. First, robots use the (x,y) coordinates of visited cells to generate $h_{s}$. Second, the action hash $h_{a}$ encodes the ``stop'' action. Third, in case a valid proof $\pi$ is generated with $h_{s}$ and $h_{a}$, robots execute the \textbf{Stop} behavior, which makes them find the center of the visited cell and stop there. After the \textbf{Stop} behavior is executed, robots cannot run other commands and they remain moveless until the end of the run.

\subsection*{S5 - IRIDIA Swarm Marketplace}
\label{sec:MarketBasedSwarm}

% Robots and CPS are going to be an important part of our future
Academic and industry leaders agree that we are currently on the verge of the fourth industrial revolution \cite{maynard2015navigating, world2016future, schwab2017fourth}, one that may be marked by the omnipresence of networks instead of stand-alone systems. More and more machines are interconnected in these networks, objects such as sensors, smart devices, factories, and so on, are gradually creating a global industrial Internet of Things (IoT) \cite{Sisini2018}. It is expected that this fourth industrial revolution will create new ways for people, machines, and organisations to be interconnected and communicate with each other by exchanging and sharing information. At the core of this new set of technologies there is a fundamental shift from centralized and human-mediated systems to decentralized and autonomous systems \cite{park2014iot}. Autonomous robots \cite{bekey2005autonomous,fahimi2009autonomous} and cyber-physical systems \cite{lee2015cyber} belong to a growing category of devices that can be programmed to perform tasks with little or no human intervention. They can vary significantly in size, functionality, mobility, dexterity, intelligence, and cost, to provide from robotic process automation to data capturing capabilities. 

% There is a new approach in the blockchain field to increase the autonomy of a process (DAO)
In the meantime, in the blockchain space, new concepts such as the Decentralized Autonomous Organization (DAO) \cite{Hsieh2018, swan2015blockchain, swan2015blockchain2} are starting to gain popularity. A DAO is an organization that is run through rules encoded as smart contracts (i.e., computer code embedded in a blockchain that directly controls the transfer of digital assets between parties under certain conditions) \cite{buterin2014next}. In these organizations, machines and people can cooperate without the need to be incorporated into traditional business identities \cite{lianos2019regulating}. Such autonomous organizations can charge users for the services they provide, in order to pay others for the resources they need. Essentially, DAOs are  entities that often need minimal or no input at all to be able to operate, and that are used to execute smart contracts and record activity on the blockchain.

% DAOs were combined but nobody created a marketplace for robot services with them
In recent years, the DAO field has also explored the combination with robotics creating a type of decentralized entity where part of the control logic resides in a smart contract while the actuation part resides in the physical world \cite{veena2015empowering}. However, there is still no example in the literature of how an institution can start offering autonomous systems services and match the needs of potential customers that might require those services on demand. With the aim to explore this path, we present the IRIDIA\footnote{IRIDIA is the name of the artificial intelligence lab of the Universit{\'e} Libre de Bruxelles, where the real-robot experiments presented in this paper took place.} Swarm Marketplace: a web-based service marketplace whose logic is coded in a smart contract and uploaded in the Ethereum blockchain. In the proposed approach, staff from IRIDIA advertises the robotics services available (number of robots, duration of service, and price). Then, customers are able to purchase these services and pay the price with their own crypto-wallets. Then they can upload  an MT with the list of operations the robots need to complete. Once the service is completed, customers get the cryptographic proof that the robots completed all the operations included in the MT, which allows them to trust the system and understand the service was not faked. Finally, customers get pictures and video footage of the final work the robots conducted. Videos are uploaded by the system to the public IRIDIA Youtube playlist together with the root of the provided MT. 
 
\begin{figure}[!t]
\centering
\includegraphics[width=0.95\linewidth]{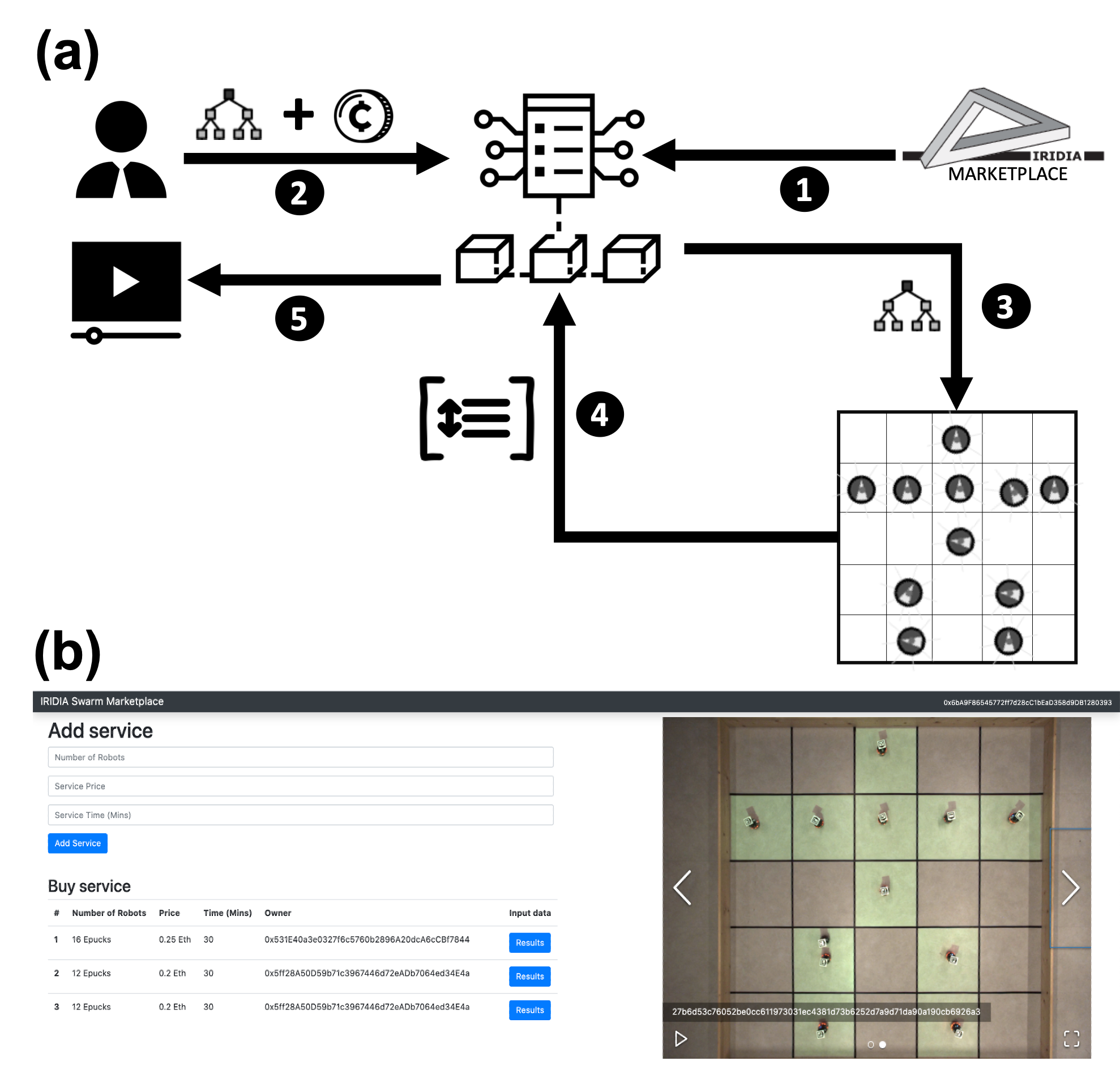}
\caption{\textbf{IRIDIA Swarm Marketplace.} (a) Information flow for the proposed application where: (1) IRIDIA staff adds services to the smart-contract by using the web interface depicted in (b). (2) Customers can purchase available services (i.e., a certain number of robots, for a time, for a price) by providing a valid MT as input and paying the price listed in cryptocurrency (i.e., Ethers). (3) Then, the smart-contract sends the MT to the robot swarm and the mission starts. (4) When the job is finished, the resultant hashes that prove that the robots did the job requested are sent to the smart-contract and displayed in the results section of the web interface. (5) The smart-contract displays the results video of the job done by the robots in a public site where the customer can retrieve it. (b) Web interface where services are added by IRIDIA staff and purchased by customers.}
\label{fig:MarketplaceWorkflow}
\end{figure}
 
% Detailed explanation of the figure
Fig.~\ref{fig:MarketplaceWorkflow} shows the general framework and main components of the IRIDIA Swarm Marketplace: a web-based marketplace for swarm robotics services built on blockchain technology. Fig.~\ref{fig:MarketplaceWorkflow}(a) depicts the main workflow of the proposed application. 
Let’s consider an example of how this market place could be used. In the planning of research activities there are periods of time in which IRIDIA’s robot swarm remains idle. When this happens, the IRIDIA staff members taking care of the robot swarm allocation to the different research activities will broadcast the robot swarm availability through a market-based website as the one depicted in Fig.~\ref{fig:MarketplaceWorkflow}(b). In this website, IRIDIA staff can log into the platform by using lab-controlled Ethereum accounts and the MetaMask interface (a popular Ethereum wallet that allows your browser to connect to the Ethereum blockchain). Then, a new service can be added where the number of robots available, the amount of time they are available for, and the price charged for their use are indicated. Once the service is broadcast, potential customers interested in using the robot swarm can purchase the service and upload their own MT in the system. After the received MT is validated, the smart-contract implementing the Swarm Marketplace sends the information to the robots so that they can start working. Robots complete the set of actions the same way it was described in the real-robot experiments of this work. In our initial demonstration example, we provide customers robots that have the maze-formation set of actions so that,  when they upload their MTs, they can choose which kind of pattern to create. Once the mission described in the uploaded MT is completed by the robots, the complete set of proofs with their correspondent $h_{a}, h_{s}$ values is sent to the smart-contract and stored there. Finally, customers can make sure their job has been completed by the robots since the values that were used as inputs for the MT have been discovered by the robots. Additionally, a video of the robots performing the mission is uploaded to the IRIDIA Youtube playlist and available for the customer. Fig.~\ref{fig:MarketplaceWorkflow}(b) shows the web interface where the buyer and seller broadcast and purchase services. A live demo of the IRIDIA Marketplace is accessible at \href{http://www.blockchainswarm.eu}{www.blockchainswarm.eu}.

%\end{document}
\end{document}